\documentclass[runningheads]{llncs}
\usepackage{color}
\usepackage{graphicx}
\graphicspath{{figures/}} 

\usepackage{hyperref}

\newcommand{\citeneeded}[1][]{{\color{blue} [citation needed]}}

\usepackage[strings]{underscore}

\usepackage{amsmath}

\usepackage{makecell}
\begin{document}

\title{Symbolic Regression for Space Applications: Differentiable Cartesian Genetic Programming Powered by Multi-objective Memetic Algorithms}

%
\titlerunning{Symbolic Regression for Space Applications}
%
\author{Marcus M\"{a}rtens\orcidID{0000-0003-1950-7111} \and
Dario Izzo\orcidID{0000-0002-9846-8423}}
\authorrunning{M. M\"{a}rtens and D. Izzo}

\institute{
European Space Agency, Noordwijk, 2201 AZ, The Netherlands\\
\email{\{marcus.maertens, dario.izzo\}@esa.int}}
\maketitle              

\begin{abstract}
Interpretable regression models are important for many application domains, as they allow experts to understand relations between variables from sparse data. Symbolic regression addresses this issue by searching the space of all possible free form equations that can be constructed from elementary algebraic functions. While explicit mathematical functions can be rediscovered this way, the determination of unknown numerical constants during search has been an often neglected issue. We propose a new multi-objective memetic algorithm that exploits a differentiable Cartesian Genetic Programming encoding to learn constants during evolutionary loops. We show that this approach is competitive or outperforms machine learned black box regression models or hand-engineered fits for two applications from space: the Mars express thermal power estimation and the determination of the age of stars by gyrochronology.
\end{abstract}

%
%

\section{Introduction}

Regression models, frequently based on machine learning concepts~\cite{fernandez2019extensive} like support vector machines~\cite{chang2011libsvm}, decision trees~\cite{loh2011classification}, random forests~\cite{breiman2001random} or neural networks~\cite{lathuiliere2019comprehensive} are common methods to predict or estimate properties of interest. However, a difficult balance needs to be found when regression arises as part of a real-world engineering problem: while complex models may provide a high accuracy, they can be opaque. For example, while certain neural network models are powerful enough to approximate any function~\cite{hornik1989multilayer}, their nonlinear and heavily parameterized layered structure is prohibitive to interpretation. For many space-related problems, a lack of interpretability is a show stopper, as predictability and robustness of the systems are of critical importance. 

Consequently, simpler models like linear regressors~\cite{weisberg2005applied} are favored in such cases, as they allow to study the behavior of the model analytically. The price of assuming a linear model is often a suboptimal accuracy if non-linear relationships are dominant in the underlying data. A human expert in the loop might instead impose certain types of functions (e.g., logarithmic, polynomial, etc.) to improve the fit, but this pragmatic approach introduces human bias and scalalibity problems.

Symbolic regression approaches aim to substitute this expert by an impartial system that will discover the best function by evolutionary recombination of basic mathematical operators, like addition, multiplication, trigonometrical functions, exponentials and similar. This allows various approaches~\cite{udrescu2020ai,schmidt2009solving,mcconaghy2011ffx} the rediscovery of explicit mathematical functions from sparse samples. In contrast, work that applies the same techniques to practical engineering problems is, putting a few noteworthy industrial exceptions~\cite{wang2019symbolic} aside, seemingly rare, raising the question why this is the case?

A potential explanation could be that numerical constants are essential to describe any reasonably noisy and complex application.
Constants are, however, rarely found in the expressions deployed to test the effectiveness of symbolic regression techniques. While those techniques excel at recombining the basic operators, they are poorly equipped to invent constants, either relying on providing the correct constants already as a fixed (sometimes mutatable) input or producing them by idiosyncratic combinations of operators.\footnote{For example, the constant $3$ might be constructed from an arbitrary input variable $x$ by evolving the expression $\frac{x}{x} + \frac{x}{x} + \frac{x}{x}$.}

In this report, we introduce a new multi-objective memetic algorithm based on differentiable Cartesian Genetic Programming (dCGP)~\cite{izzo2017differentiable,izzo2020dcgp} that is capable of finding interpretable mathematical expressions for two different real world problems coming from the space domain without the need for specific expert knowledge. We show that by exploiting the differentiability of a dCGP expression, the best fitting constants can be learned directly. The evolved expressions are shown to be either competitive or outperforming state of the art reference models for the tasks under consideration, including machine learned regression models. Due to its multi-objective nature, a decision maker is able to balance interpretability (i.e. expression complexity) with model accuracy. 

Our report is structured as follows: Section~\ref{sec:related} provides some context of previous symbolic regression works and relates them to ours. Section~\ref{sec:symbolic} briefly explains dCGP and introduces our memetic and multi-objective algorithm. Section~\ref{sec:case1} and Section~\ref{sec:case2} describe two case studies, one from the Mars express mission and one from star dating, to which our algorithm is applied and evaluated. We conclude our work in Section~\ref{sec:conclusions}.

\section{Related Work}
\label{sec:related}

Early works on symbolic regression were inspired by the Genetic Programming approach of Koza~\cite{koza1994genetic} and deployed encodings based on arithmetic trees or grammars to evolve mathematical expressions. The encoding of our chromosomes is based on an extension of Cartesian Genetic Programming (CGP), which was originally developed by Miller~\cite{miller2011cartesian}.
A comprehensive overview about recent genetic symbolic regression techniques and their applications is given by Wang et. al~\cite{wang2019symbolic}. Most influential is the seminal work of Michael Schmidt and Hod Lipson~\cite{schmidt2009distilling} who extracted conservation laws from observing physical systems like the double pendulum. Their evolutionary technique is available in the commercial ``Eureqa'' software. Eureqa uses an evolutionary technique to determine the numerical value of constants.

There also exist approaches (the deterministic FFX algorithm~\cite{mcconaghy2011ffx}) that do not rely on genetic programming. Most recently, Udrescu and Tegmark~\cite{udrescu2020ai} introduced the ``AI Feynman algorithm'' which relies on a series of regressions techniques, brute-forcing and neural networks instead of evolutionary approaches. However, AI Feynman requires constants to be provided as input a priori or must otherwise construct them by operator recombination.


\section{Symbolic Regression with dCGP}
\label{sec:symbolic}

\subsection{dCPG}

Differentiable Cartesian Genetic Programming (dCGP) is a recent development in the field of Genetic Programming (GP) adding to genetic programs their any-order differential representation to be used in learning tasks~\cite{izzo2020dcgp}.
The use of low-order differentials, i.e. gradients, to learn parameters in Genetic Programs is rare in previous works and can be found, for example, in \cite{topchy2001faster} and  \cite{emigdio2015local}. 
In past research, attempts to use genetic operators such as crossover and mutation \cite{howard1995ga} as well as meta-heuristics such as simulated annealing \cite{bitch} were made to adapt such parameters. 
It is clear how access to the model's differential information, when available, may bring substantial gains if used during learning as proved, for example, by the enormous success of stochastic gradient descent in machine learning and in particular in learning deep neural network parameters. Key to the successful use of the differential information is its availability at a low computational cost, a feature that for first order derivatives, i.e. the gradient, is guaranteed by the backward automated differentiation technique, while for higher order derivatives is more difficult to obtain. 
Thanks to the built-in efficient implementation of the algebra of Taylor truncated polynomials, evaluating a dCGP program allows seamless access to gradients and Hessians (as well as higher order differential information).
In the following, we describe how our new Multi Objective Memetic Evolutionary Strategy (MOMES) leverages the loss gradient and its Hessian to evolve the dCGP program and learn the ephemeral constants simultaneously.

\subsection{Multi Objective Memetic Evolutionary Strategy (MOMES)}
Let us denote with $\boldsymbol \eta = (\boldsymbol \eta_u, \mathbf c)$ a generic chromosome encoding the symbolic expression $\hat y = f_{\boldsymbol \eta_u}(\mathbf x, \mathbf c)$. We recognize in $\boldsymbol \eta $ two distinct parts: $\boldsymbol \eta_u$ is the integer part of the chromosome making use of the classical Cartesian Genetic Programming \cite{miller2011cartesian} encoding. Thus, $\boldsymbol \eta_u$ represents the mathematical form of $\mathbf f$, while $\mathbf c$ represents the actual values of the ephemeral constants in a continuous direct encoding. We may then introduce the gradient and Hessian as:
$$
\begin{array}{l}
\nabla \hat y = [\frac{\partial \hat y}{\partial c_1}, ..., \frac{\partial \hat y}{\partial c_n}]
\hspace{1.5cm}
\nabla^2 \hat y = \begin{bmatrix} 
   \frac{\partial \hat y^2}{\partial c_1\partial c_1} & \dots &  \frac{\partial \hat y^2}{\partial c_1\partial c_n} \\
    \vdots & \ddots & \\
    \frac{\partial \hat y^2}{\partial c_n\partial c_1} &        & \frac{\partial \hat y^2}{\partial c_n\partial c_n}
    \end{bmatrix}
\end{array}
$$
Assuming to work on labelled data denoted by $\mathcal D = (\mathbf x_i, y_i)$, $i=1..N$ we may thus compute, for any chromosome $\boldsymbol \eta$, the mean square error (MSE) loss:
$$
\ell = \sum_i (y_i - \hat y_i)^2
$$
its full Hessian ${\mathbf H}$ and its full gradient ${\mathbf G}$. We also define the complexity $\mu$ of the chromosome $\boldsymbol \eta$ as the number of active nodes in its CGP. 
Let us now introduce the concept of active constants $\tilde {\mathbf c}$ as the selection from $\mathbf c$ of all components with nonzero entries in the gradient ${\mathbf G}$. Zero entries correspond, most of the time, to ephemeral constants that are not expressed by the corresponding CGP and are thus inactive. Similarly, we call $\tilde{\mathbf H}$ and $\tilde{\mathbf G}$, respectively, the loss active Hessian and the loss active gradient when considering only the active constants. Following these notations and definitions, we may now describe the MOMES algorithm.

MOMES is an evolutionary strategy where all members in a population of $NP$ individuals, each represented by a chromosome $\boldsymbol \eta_i$, $i=1..NP$, undergo a mutation step at each generation, acting on all the genes (active and inactive) in $\boldsymbol \eta_u$. The mutated chromosome is then subject to a lifelong learning process consisting of one single step of Newton's method and acting on the (active) continuous part $\tilde {\mathbf c}$ of the chromosome:
$$
\tilde {\mathbf c}_{new} = \tilde {\mathbf c}_{old} - \tilde{\mathbf H}^{-1}\tilde{\mathbf G}
$$
The new individual fitness is then evaluated and, to preserve diversity, the chromosome is added to the $NP$ parents only if its fitness is not already present in the candidate pool.
Non-dominated sorting over the objectives $(\ell, \mu)$ is then applied to the candidate pool selecting $NP$ individuals to be carried over to the new generation. The exact implementation details of MOMES are documented in the open source project dCGP~\cite{izzo2020dcgp}.

The design of MOMES required the use of a few ideas that were the results of experiments and iterations that are worth discussing briefly at this point. Firstly, the mutation of the CGP expression encoded in $\boldsymbol \eta_u$ acts randomly on all genes, also the ones that are not expressed by the resulting expression. This type of mutation, already noted as beneficial when used in the standard evolutionary strategy technique used for CGP~\cite{miller2011cartesian}, has an added benefit here as it allows for a mutated chromosome to express the same formula as the parent. In these cases, the subsequent lifelong learning ensures the possibility to apply further iterations of the Newton's method to the same expression, and the mechanism will thus ensure that, eventually, a full local descent and not only one single iteration is made over the most promising individuals. Additionally, this simple trick acts as a necessary diversity preservation mechanism as otherwise the very same expression (possibly having a different complexity as counted by the number of active nodes in the CGP) would flood the non-dominated front and lead to a degraded convergence.
Secondly, the choice of using a single step of a second order method for the memetic part of MOMES was found to be most effective when coupled with MSE loss. It has been noted (see \cite{izzo2017differentiable}) that, whenever constants appear linearly in the expressed CGP, one single step of a second order method is enough to get to the optimal choice if the loss is expressed as MSE. This mechanism adds a small, but beneficial bias towards simple expressions during evolution, since these are evaluated with optimal constants values at each generation, while more complex appearances of constants would require more generations to reach their optimal values.
Lastly, the introduction of active constants, Hessians and gradients allows the Hessian matrix to admit an inverse and thus enable an effective lifelong learning.

\section{Case 1: Mars Express Thermal Data}
\label{sec:case1}

\subsection{Background and Dataset}
\label{subsec:mexdataset}

\begin{figure}[h]
\includegraphics[width=\textwidth]{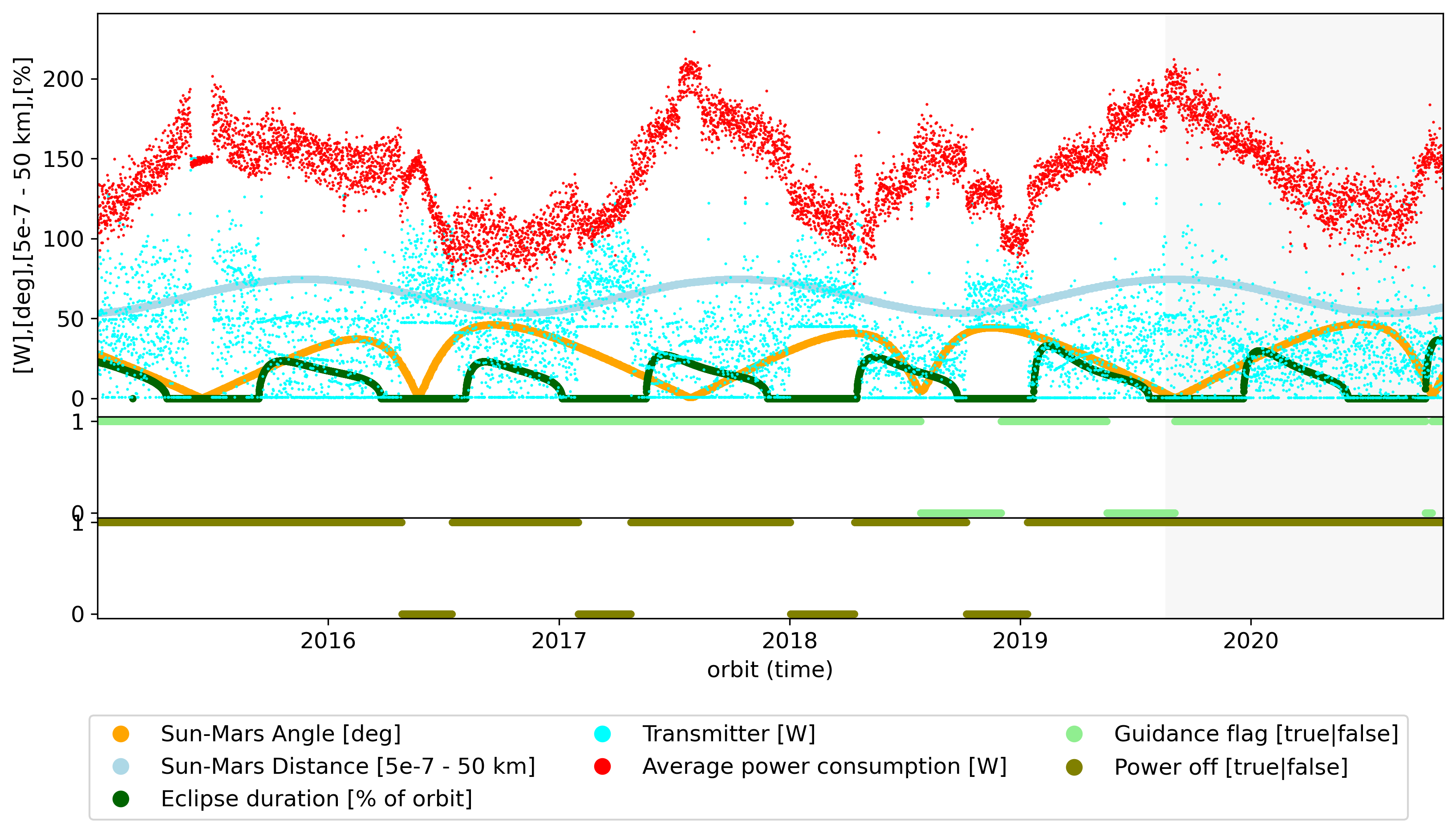}
\caption{MEX data, including orbital and operative parameters. Goal is to predict the average power consumption from thermal contributors, compare Equation~\ref{eq:thcontrib}.}
\label{fig:mex1original}
\end{figure}

Mars Express (MEX) is a spacecraft of the European Space Agency (ESA) that has been monitoring the ``red planet'' continuously since 2004. 
To remain operable in the challenging environment of space, the probe is equipped with a thermal regulation system to which a certain power has to be allocated. 
An accurate power budget is essential for the mission, as underestimates might lead to malfunctions and overestimates will waste energy that should have been allocated to scientific tasks instead.
The MEX data for this work~\cite{maertens2022mars} is split in a training set MEX1 and a test set MEX2. MEX1 consists of 5795 completed orbits in the time frame from 01.01.2015 to 18.08.2019 and MEX2 to 1507 orbits from 18.08.2019 to 31.10.2020. Measurements of potential relevance for the prediction of the thermal power budget have been aggregated for each orbit into several thermal contributors. In particular, the task is to find the thermal power demand $P_{th}$ as a function
\begin{equation}
\label{eq:thcontrib}
P_{th}(LVAH, SH, D_{ecl}, TX, FO, NS).
\end{equation}
A reference model based on linear regression
\begin{equation}
\label{eq:refmodel}
P_{th} = c \cdot LVAH + d \cdot SH + e \cdot D_{ecl} + f \cdot TX + g \cdot FO + h \cdot NS + i
\end{equation}

was fitted to the MEX1 data. Table~\ref{tab:mex} gives a short description of each thermal contributor and the value of the fitted coefficients. Figure~\ref{fig:mex1original} shows the MEX data in relation to orbital conditions and actual power consumption. The evaluation metric for all MEX data is the Root Mean Square Error (RMSE).

\begin{table}[h!]
\centering
\begin{tabular}{|c|c|l|c|}
\hline 
\thead{thermal\\ contri-\\ butor} & \thead{description} &  \thead{related \\ coefficient} & \thead{dCGP \\ variable} \\ 
\hline 
$LVAH$ & Launch Vehicle Adapter heating & $c = -7.666 \cdot 10^1$ & $x_0$ \\ 
\hline 
$SH$ & solar heating & $d = -1.764 \cdot 10^{-1}$ & $x_1$ \\ 
\hline 
$D_{ecl}$ & heating due to eclipses & $e = -3.387 \cdot 10^{-3}$ & $x_2$ \\ 
\hline 
$TX$ & transmitter activities & $f = -6.898 \cdot 10^0$ & $x_3$ \\ 
\hline 
$FO$ & flag indicating power off & $g = +1.107 \cdot 10^1$ & $x_4$ \\ 
\hline 
$NS$ & guidance flag & $h = +6.820 \cdot 10^0$ & $x_5$ \\ 
\hline 
- & (baseline) & $i = +2.267 \cdot 10^2$ &  \\ 
\hline 
\end{tabular} 
\caption{The reference model THP for MEX data is a linear regression model with the highlighted coefficients (compare Equation~\ref{eq:refmodel}). Last column shows corresponding variables for dCGP expressions.}
\label{tab:mex}
\end{table}

\subsection{Results}
\label{subsec:mexresults}

For this experiment, MEX1 data is used for training and MEX2 data for additional testing. Our features are the thermal contributors of the reference THP model (see Table~\ref{tab:mex}), with two differences: $SH$ and $D_{ecl}$ are standardized by subtracting the mean and dividing by the standard deviation of the training set. The determination of optimal hyperparameters for dCGP and MOMES is not within the scope of this work, as it would require us to thoroughly explain, characterize and statistically analyze each of them. In practice, evolution converges fast enough to perform grid searches over some reasonable parameter guesses that can be further calibrated over a few iterations of the optimization pipeline. Table~\ref{tab:mexhyper} reports the hyperparameters that we found to be effective enough to make our point with this experiment.

\begin{figure}[h!]
\includegraphics[width=\textwidth]{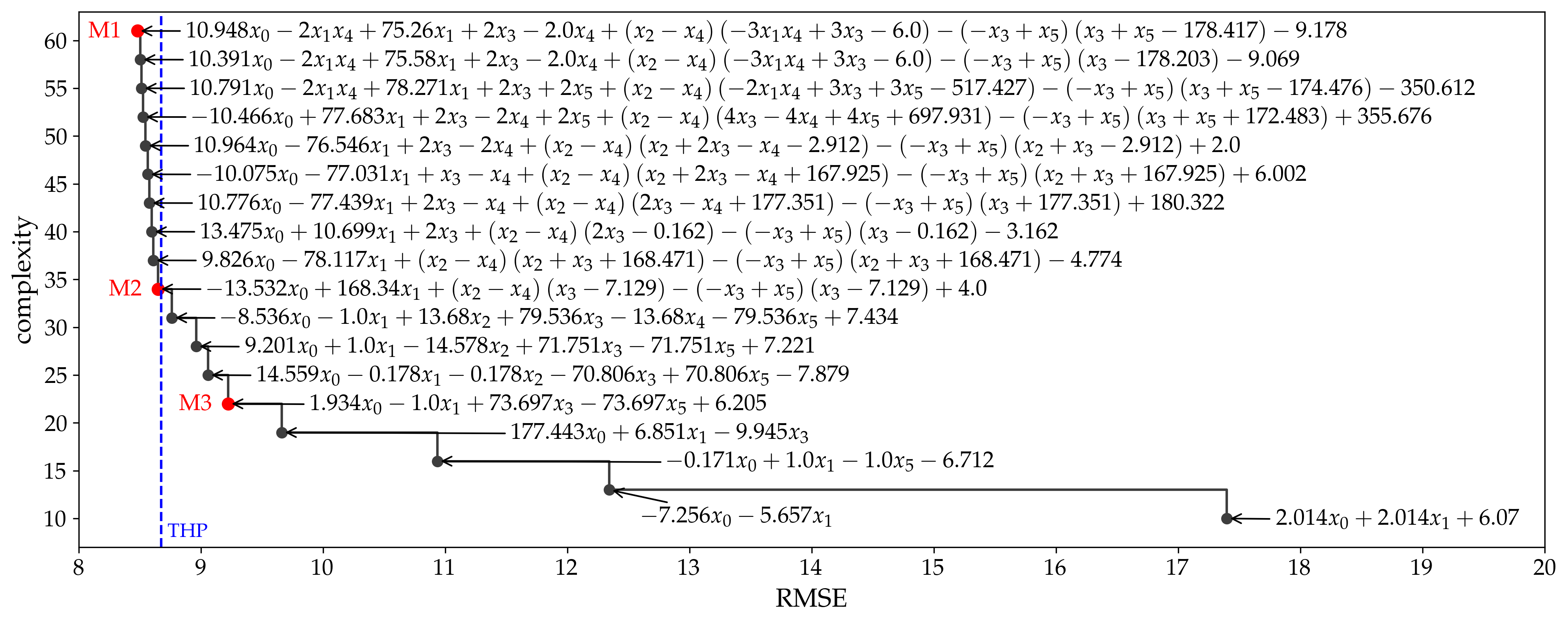}
\includegraphics[width=\textwidth]{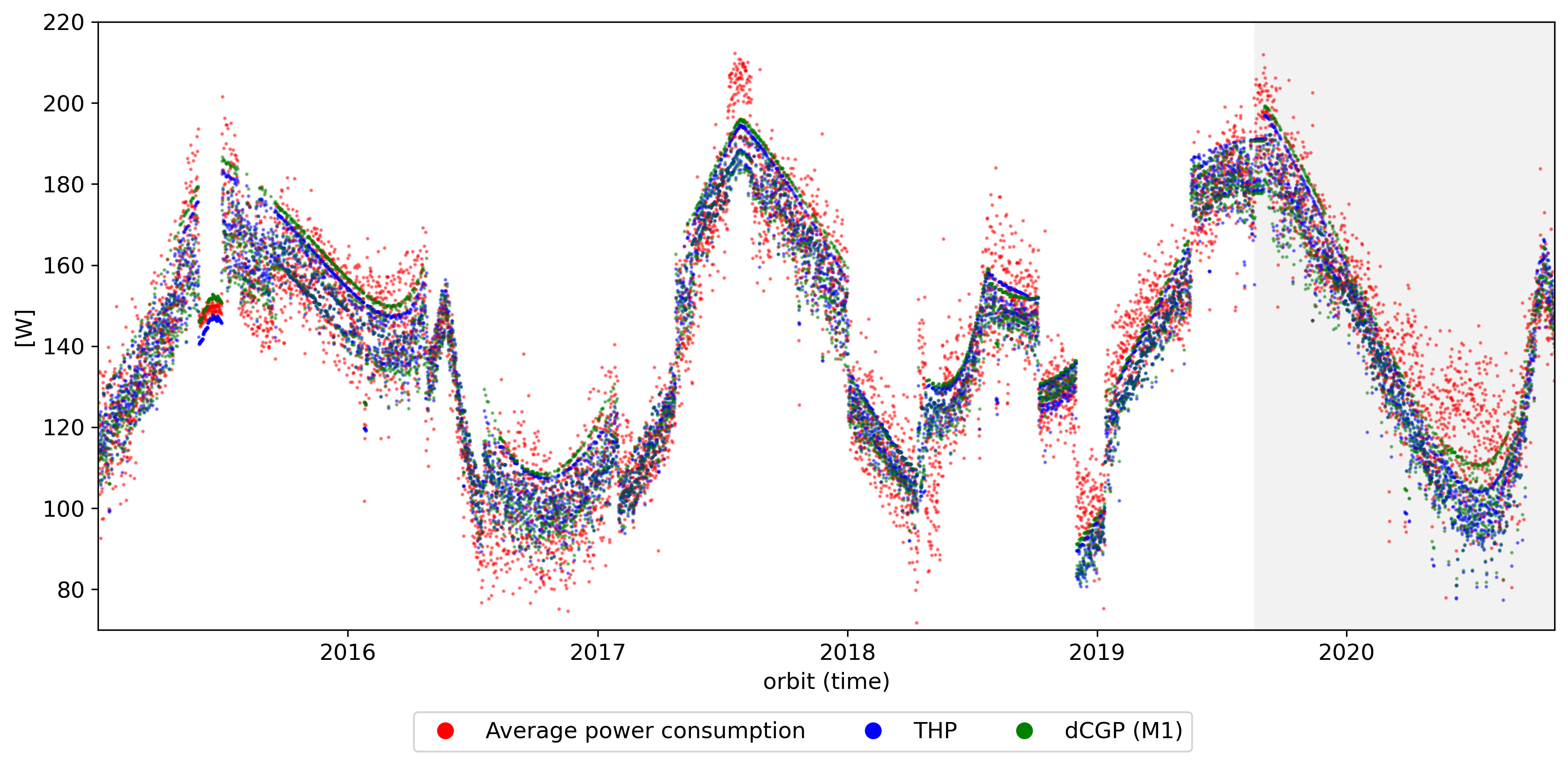}
\caption{\textbf{Top:} Non-dominated front found by MOMES, 18 out of 40 individuals, MEX1 data. Blue line is the reference model THP. The expressions M1, M2 and M3 are highlighted for further analysis. \textbf{Bottom:} Predictions of THP and dCGP expression M1 in direct comparison with target power consumption. Shaded region is test data MEX2.}
\label{fig:mexpareto}
\end{figure}

\begin{table}
\centering
\begin{tabular}{|c|c|}
\hline 
\thead{parameter} & \thead{value} \\ 
\hline 
basic operators & $+, -, \times, \div, \log$ \\ 
\hline 
rows & 2 \\ 
\hline 
columns & 20 \\ 
\hline 
levels back & 20 \\ 
\hline 
maximum number of constants & 5 \\
\hline 
maximum mutations & 4 \\ 
\hline 
generations & 50000 \\ 
\hline 
population size & 40 \\ 
\hline 
\end{tabular} 
\caption{Hyperparameters deployed for dCGP and MOMES for the MEX1 dataset.}
\label{tab:mexhyper}
\end{table}

As common with evolutionary techniques, we multi-start MOMES 200 times to account for its stochasticity. Out of those 200 results, we select the one that has the strongest extreme point (i.e. lowest RMSE) and show the complete Pareto front in Figure~\ref{fig:mexpareto}. For further comparison with the THP reference model, we select three different formulas: M1 is at the extreme point, M2 is the expression with minimum complexity that is still above the reference and M3 is a low complexity expression. Table~\ref{tab:mexresults} shows the corresponding RMSE and the average over- and underestimations.

\begin{table}[h]
\centering
\rotatebox[origin=c]{90}{MEX1}
\begin{tabular}{|c||c|c|c|c|}
\hline 
\thead{Model} & \thead{RMSE} & \thead{complx.} & \thead{avg. over-\\estimate} &  \thead{avg. under-\\estimate} \\ 
\hline THP & 8.672 & - & 6.665 & 6.774 \\
\hline M1 & 8.478 & 61 & 6.657 & 6.442 \\
\hline M2 & 8.647 & 34 & 6.826 & 6.619 \\
\hline M3 & 9.223 & 22 & 7.184 & 7.232 \\
\hline
\end{tabular}\\
\rotatebox[origin=c]{90}{MEX2}
\begin{tabular}{|c||c|c|c|c|}
\hline 
\thead{Model} & \thead{RMSE} & \thead{complx.} & \thead{avg. over-\\estimate} &  \thead{avg. under-\\estimate} \\ 
\hline THP & 12.454 & - & 10.684 & 4.549 \\
\hline M1 & 11.390 & 61 & 9.753 & 4.461 \\
\hline M2 & 12.828 & 34 & 10.924 & 4.998 \\
\hline M3 & 14.991 & 22 & 12.928 & 6.745 \\
\hline
\end{tabular} 
\caption{RMSE for 3 selected expression from Figure~\ref{fig:mexpareto}, complexities and values for average over- and underestimation of thermal power consumption. THP is the linear regression reference model.}
\label{tab:mexresults}
\end{table}

We observe that the RMSE for MEX2 is considerably higher than for MEX1 for each of the models, with Expression M1 generalizing slightly better to the unseen conditions of MEX2 than THP. We show the actual predictions of both, THP and Expression M1 in Figure~\ref{fig:mexpareto}, which highlights the differences between the models. From a mathematical point of view, M1 includes several products of variables, resulting in a degree 2 polynomial. Despite having access to the division and logarithm operator, evolution did not favor them in the population. However, other non-dominated fronts from the 200 runs (not shown) included sometimes division, constructing rational functions of similar RMSE. M3 is a linear model without the thermal contributors $x_2$ and $x_3$ that is worse than THP but could provide an interesting alternative if certain measurements would not be readily available. While some of the numerical constants are similar across different expression, it is clear that the memetic algorithm is optimizing them for each expression independently to further reduce the error. Furthermore, the maximum of 5 constants is not (always) expressed, highlighting that MOMES is selective and parsimonious. In comparison, the simple linear regression model THP relies already on 7 explicit constants.

\section{Case 2: Star Dating Gyrochronology}
\label{sec:case2}

\subsection{Background and Dataset}
\label{subsec:stardataset}

This dataset was released by Moya et al.~\cite{moya2021ai} as a benchmark for AI research on star dating. Many physical models for astronomical problems (e.g. the search for habitable exoplanets) depend on the age of a star, which is however not directly measurable. Consequently, the age of a star needs to be inferred from observable features. The public benchmark describes 1464 stars whose accurate ages have been determined by asteroseismology and association to star clusters, providing for each star the stellar features described in Table~\ref{tab:starfeatures}.

In this context an approach termed ``gyrochronology'' is of particular interest, which hypothesizes a functional relationship between star age, stellar mass and rotational period. Some empirical relations (assuming a fixed functional form) have been proposed~\cite{angus2015calibrating} but the accuracy of gyrochronology remains controversial, with some works suggesting that linear relationships might not be sufficient at all or just applicable for stars of a certain age group~\cite{barnes2010simple}. Consequently, the authors of the dataset propose to study star dating directly as regression problem using supervised machine learning models.

\begin{table}
\centering
\begin{tabular}{|c|c|c|}
\hline 
\thead{Feature} & \thead{Description} &  \thead{dCGP \\ variable} \\ 
\hline 
$M$ & Stellar Mass & $x_0$ \\ 
\hline 
$R$ & Stellar Radius & $x_1$ \\ 
\hline 
$T_{eff}$ & Stellar effective temperature & $x_2$ \\ 
\hline 
$L$ & Luminosity & $x_3$ \\ 
\hline 
$[Fe/H]$ & Metallicity & $x_4$ \\ 
\hline 
$\log g$ & Logarithm of surface gravity & $x_5$ \\ 
\hline 
$P_{rot}$ & Stellar rotational period & $x_6$ \\ 
\hline 
\end{tabular} 
\caption{Features available for every star in the dataset.}
\label{tab:starfeatures}
\end{table}

Several reference models including linear regression, decision tree regressor, random forest regressor, support vector regressor, Gaussian process, kNN, neural networks and ensembles are provided within the benchmark. Additionally, four different splits of the data for training and test are available:
\begin{itemize}
\item \textbf{A} a random 80/20 split
\item \textbf{B1} stars of age $[0.4, 4.2]$ Gyr are used for training, ages $[4.2, 13.8]$ for testing
\item \textbf{B2} stars dated by cluster belonging used for training, remainder for testing
\item \textbf{C} independent control with 32 new stars for testing, including our own sun
\end{itemize}

The evaluation metric for all gyrochronology models is the Mean Absolute Error (MAE). Additionally, the dataset provides error bounds on the age of the stars allowing to define a precision metric by the fraction of star age predictions that fall within the corresponding confidence interval.

\subsection{Results}
\label{subsec:starresults}

For this experiment, we deployed MOMES on all training sets of the given benchmarks and re-evaluated the found expressions on the test sets. We selected the 6 features as described in Table~\ref{tab:starfeatures} and scaled them by dividing each of them by the standard deviation of the corresponding training set. 
Similar to Section~\ref{sec:case1}, a grid-search was used to find suitable hyperparameters for dCGP and MOMES, reported in Table~\ref{tab:starhyper}.

\begin{table}
\centering
\begin{tabular}{|c|c|}
\hline 
\thead{parameter} & \thead{value} \\ 
\hline 
basic operators & $+, -, \times, \div, \log, \sin$ \\ 
\hline 
rows & 2 \\ 
\hline 
columns & 16 \\ 
\hline 
levels back & 16 \\ 
\hline 
number of constants & 5 \\
\hline 
maximum mutations & 6 \\ 
\hline 
generations & 500000 \\ 
\hline 
population size & 40 \\ 
\hline 
\end{tabular} 
\caption{Hyperparameters deployed for dCGP and MOMES for star dating.}
\label{tab:starhyper}
\end{table}

\begin{figure}
\includegraphics[width=\textwidth]{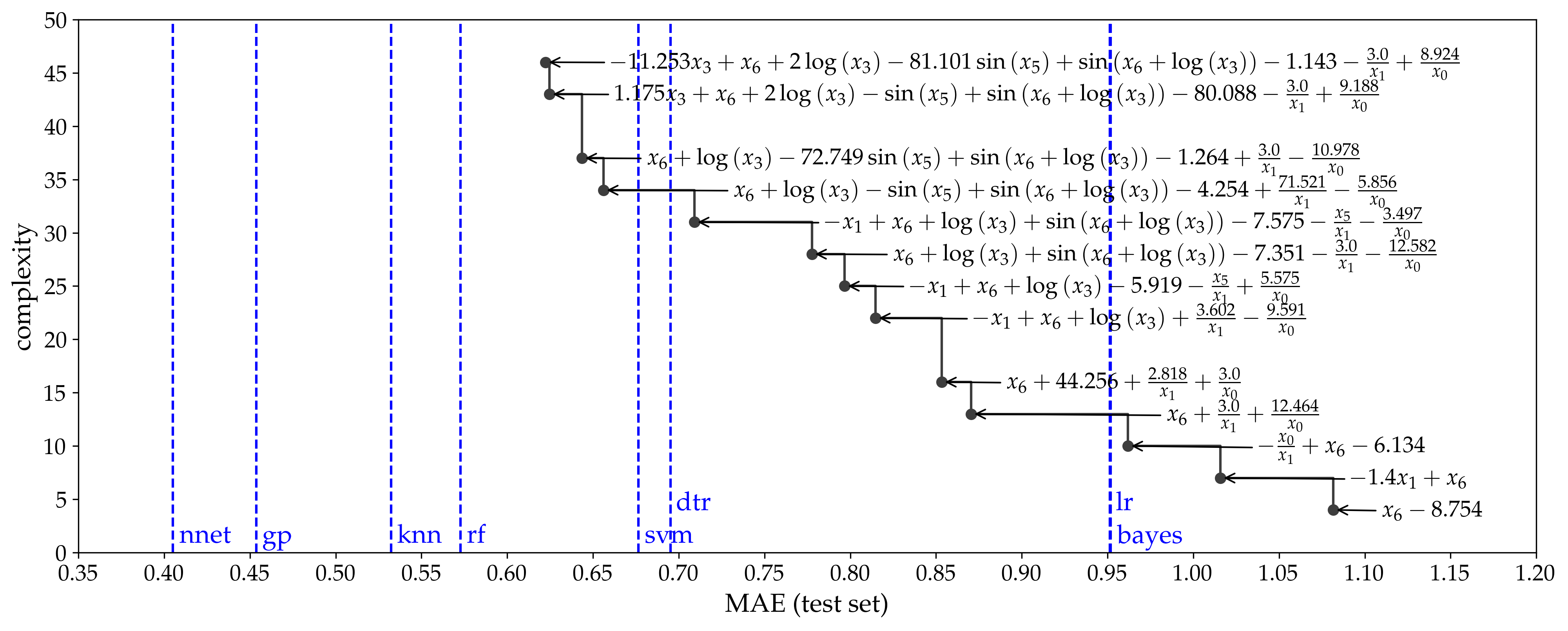}\\
\includegraphics[width=\textwidth]{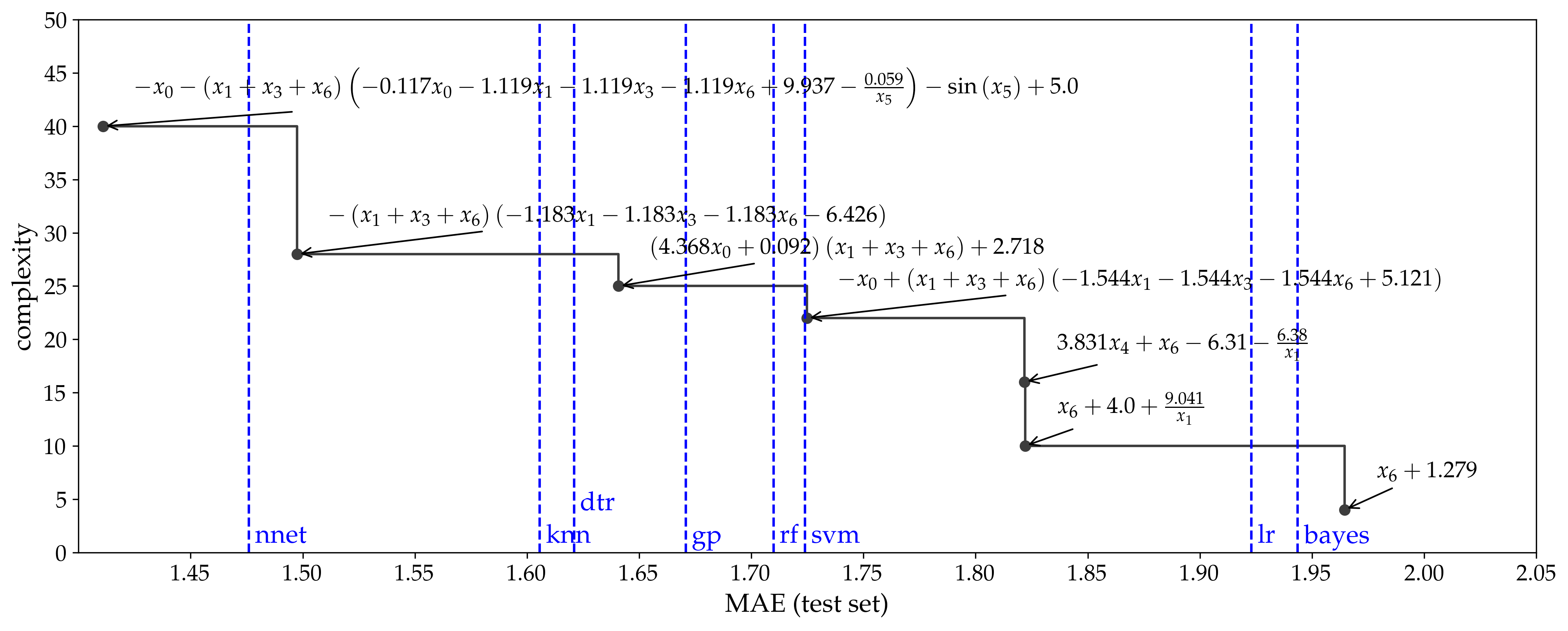}
\caption{Non-dominated front found by MOMES for \textbf{Benchmark A (top)} and \textbf{B1 (bottom)}. Shown is the non-dominated front after the complete population was re-evaluated on the test-set of its corresponding benchmark. Blue lines represent reference machine learned models as described by Moya et al.~\cite{moya2021ai}}
\label{fig:starpareto}
\end{figure}

A multi-start setup of 400 independent runs (100 for each benchmark) has been deployed to our servers and completed in about 12 hours.
We selected the run that resulted in the lowest MAE (extreme point) on the training set for further analysis. Table~\ref{tab:starresults1} shows the MAE and the precision of the lowest MAE expression found in the non-dominated front of MOMES in comparison with the reference models. To exemplify the diversity and complexity trade-offs in the found expressions, we show the approximated Pareto fronts in Figure~\ref{fig:starpareto} for Benchmark A and B1 (for reasons of space).

Interpreting the results, the dCGP based models are competitive with some of the machine learning models for benchmark A, but have a higher error than neural networks, Gaussian processes, kNN and random forests. Benchmark B1 represents a domain gap study, which is (without the deployment of additional techniques or assumptions) hard for most machine learning models to bridge. Since dCGP looks to capture the physical relations between the different features in algebraic expressions, it leads itself to better generalization.

Investigating the Pareto front of dCGP on the test-set for B2, the simple expression $x_6 - \sin (x_1 - x_4) + 6.987$ could be shown to outperforms all other machine learning models by some margin. On a closer look however, it turned out that the correlation between the training error and test error in this benchmark was rather low when looking at the entirety of our MOMES runs. Although this particular result for B2 has to be taken with a grain of salt, it nevertheless demonstrates the existence of surprisingly simple expressions that provide an accurate explanation of the data. Thanks to the interpretability of dCGP expressions, these type of insights may uncover potential issues related to overfitting of machine learning models or issues related to the data split that would go unnoticed otherwise. Lastly, on benchmark C the best dCGP expressions achieves a low MAE, only surpassed by the Gaussian process model.

\setlength{\tabcolsep}{0.6em}

\begin{table}
\centering
\rotatebox[origin=c]{90}{Precision~~~~~~~~~~~~MAE~~}
\begin{tabular}{|c||c|c|c|c|c|c|c|c|c|}
\hline 
\thead{} & \thead{dcgp} & \thead{nnet} & \thead{lr} &  \thead{dtr} & \thead{rf} & \thead{svm} &  \thead{bayes} & \thead{knn} & \thead{gp} \\ 
\hline 
A & 0.62 & 0.41 & 0.95 & 0.70 & 0.57 & 0.68 & 0.95 & 0.53 & 0.45 \\
B1 & 1.41 & 1.48 & 1.92 & 1.62 & 1.71 & 1.72 & 1.94 & 1.61 & 1.67 \\
B2 & 1.39 & 1.85 & 1.56 & 1.90 & 1.85 & 1.51 & 1.48 & 1.84 & 1.62 \\
C & 0.97 & 1.10 & 1.30 & 2.11 & 1.01 & 1.28 & 1.30 & 1.73 & 0.84 \\
\hline
\hline 
\thead{} & \thead{dcgp} & \thead{nnet} & \thead{lr} &  \thead{dtr} & \thead{rf} & \thead{svm} &  \thead{bayes} & \thead{knn} & \thead{gp} \\ 
\hline 
A & 50.39 & 73.23 & 35.43 & 63.78 & 60.63 & 46.46 & 32.28 & 77.95 & 62.20 \\
B1 & 51.97 & 59.84 & 29.92 & 51.18 & 48.03 & 37.79 & 28.35 & 55.91 & 51.97 \\
B2 & 42.26 & 28.45 & 30.96 & 21.34 & 25.94 & 33.05 & 29.71 & 25.52 & 30.96 \\
C & 15.63 & 34.37 & 21.87 & 9.37 & 21.87 & 12.50 & 21.87 & 15.62 & 40.62 \\
\hline
\end{tabular} 
\caption{MAE and precision on test set for all star dating benchmarks. Each model was trained for each benchmark separately. The dCGP expressions used are always the ones that are at the extreme end of the MAE objective of the non-dominated front created by MOMES.}
\label{tab:starresults1}
\end{table}


\section{Conclusions}
\label{sec:conclusions}

Following our experiments on the Mars express and star dating datasets, we have demonstrated that MOMES is capable of finding algebraic expressions that explain unseen data and have the capability to generalize well to it. The fact that the numerical constants for each expression are learned at each step of evolution guides the memetic algorithm towards expressions that are both accurate and of low complexity. The expressions with the lowest error notably contain products of sub-expressions or non-linear operators like the $\sin$-function and are thus more expressive than linear regressors while still amenable to mathematical analysis. Consequently, no expert is required to make potentially biased guesses in finding non-linear expressions. The task that still remains for the expert is the engineering and scaling of features (e.g. the thermal contributors in case of MEX) to obtain the best possible results, as well as the determination of suitable basic operators and efficient hyperparameters.

As shown in the case of benchmark B2 for the star dating case study, deploying symbolic regression in general is beneficial as sometimes simple expressions appear to be more general than complex black box models. Given the practical benefits of interpretability that comes with algebraic expressions together with the low requirements for their inference make algorithms like MOMES valuable methods of knowledge discovery. In particular, applications with sparse data in extreme environments like space are likely to benefit from it.

\subsubsection*{Acknowledgements}
The authors are grateful to Thomas Dreßler, who helped making the MEX data public and explained it to us. Similarly, the authors want to thank Roberto J. López-Sastre for providing support and insight into the gyrochronology data.

%
%

\bibliographystyle{splncs04}
\bibliography{references}

\end{document}